\pdfoutput=1

\documentclass[11pt]{article}

\usepackage[]{ACL2023}
\usepackage{times}
\usepackage{latexsym}

\usepackage[T1]{fontenc}

\usepackage[utf8]{inputenc}

\usepackage{microtype}

\usepackage{inconsolata}
\usepackage{subfigure}
\usepackage{caption}
\usepackage{graphicx}
\usepackage{setspace}%
\usepackage{amsmath}
\usepackage{amssymb}
\usepackage{color}
\usepackage{multirow}
\usepackage{booktabs}
\title{Causal Interventions-based Few-Shot Named Entity Recognition}

\author{Zhen Yang \\
  University of South China  \\
    \texttt{uscyz094@gmail.com} \\\And
  Yongbin Liu \\
  University of South China \\  
  \texttt{yongbinliu03@gmail.com} \\\And
  Chunping Ouyang \\
  University of South China \\  
  \texttt{ouyangcp@126.com} \\}

\begin{document}
\maketitle
\begin{abstract}
Few-shot named entity recognition (NER) systems aims at recognizing new classes of entities based on a few labeled samples. A significant challenge in the few-shot regime is prone to overfitting than the tasks with abundant samples. The heavy overfitting in few-shot learning is mainly led by spurious correlation caused by the few samples selection bias. To alleviate the problem of the spurious correlation in the few-shot NER, in this paper, we propose a causal intervention-based few-shot NER method. Based on the prototypical network, the method intervenes in the context and prototype via backdoor adjustment during training. In particular, intervening in the context of the one-shot scenario is very difficult, so we intervene in the prototype via incremental learning, which can also avoid catastrophic forgetting. Our experiments on different benchmarks show that our approach achieves new state-of-the-art results (achieving
up to 29\% absolute improvement and 12\% on
average for all tasks).
\end{abstract}

\section{Introduction}

As a fundamental task in information extraction, named entity recognition (NER) aims at locating and classifying named entities from unstructured text. Many methods \cite{chiu2016named, ma2016end, lample2016neural, peters2017semi} have been used to achieve efficient results in named entity recognition. 

In practical applications, due to the difficulty of label collection and the high expensiveness of manual labeling, few-shot named entity recognition was proposed and has received wide attention. Many studies on few-shot NER have appeared in recent years. Current approaches are mainly based on metric learning models, in particular, on prototypical networks. The prototypical networks method calculates a prototype representation \cite{snell2017prototypical} for each class and  assigns the labels to each query sample according to the distance between the sample and the prototype of each class \cite{fritzler2019few, yang2020simple, hou2020few}. In addition, the span-based methods \cite{wang2021enhanced, yu2021few, ma2022decomposed} also emerge to be able to help few-shot named entity identification to discover the boundaries of entities better.

\begin{figure}
  \centering
  \includegraphics[width=0.45\textwidth,height=0.4\textwidth]{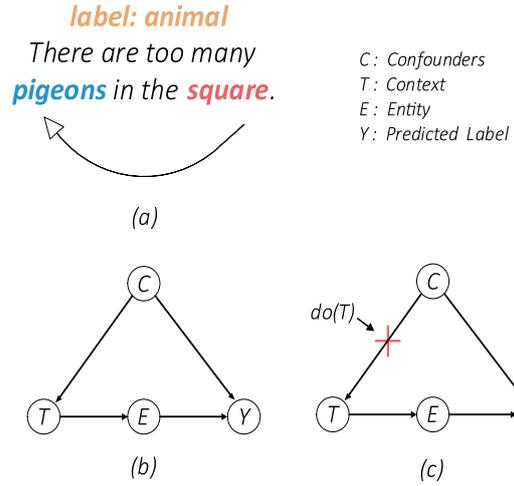}
  \caption{ (a) An example of spurious correlation. \emph{Pigeons} are easily associated with \emph{square}, but not all animals are associated with \emph{square}. (b) Causal graph of the example. Contexts \emph{T} ( such as the \emph{square}),  entities \emph{E} (such as \emph{pigeons}), and class label \emph{Y} (such as the \emph{animal} class type), \emph{C} is  confounding factors brought by samples selection bias in the few-shot task. (c) Causal graph after \emph{do}-operation}\label{1}
\end{figure}

However, these methods ignore the overfitting problem caused by spurious correlation in few-shot tasks. The spurious correlation issue is
 less severe in tasks with large samples. But due to the few samples selection bias, this issue must be addressed in the few samples tasks. As Figure 1(a) shows, in an example of the spurious correlation, the \emph{pigeons}  belong to the class label \emph{animal}. In the few-shot NER task, the \emph{animal} is easily associated with the \emph{square}. But the \emph{square} is not positively related to all animal entities. This way is a spurious correlation established by the few samples. 

All the problems can be due to the overfitting of the model to the confounding factors in very few samples. Essentially, the few samples selection bias is just a confounder that misleads the few-shot NER model to learn spurious correlations between contexts and labels, e.g., between the context \emph{square} and the ground-truth label \emph{animal} in Figure 1(a). More specifically, the confounder is prone to help for a positive association between the contexts and some entity that can be formalized as $P(E\big|T)$, e.g., When the context \emph{square} is encountered, it increases the likelihood that it is a \emph{pigeon}. The $P(E\big|T)$ based on the confounder misleads to associate the non-causal but positive correlated context \emph{T} to class label \emph{Y}, $P(Y\big|E,T)$, e.g., the context \emph{square} is wrongly regarded as the stable and intrinsic feature of the class label \emph{animal}, practically it is a non-causal feature in the class animal. 

Existing metric-based methods do not eliminate the potential confounding factors in the few samples, especially the few samples selection bias. 
To deal with the problems mentioned above, we review the few-shot named entity recognition from a causal viewpoint. Figure 1(b) shows the corresponding causal graph. We formulate the causalities among entity \emph{E}, relevant context \emph{T}, class label \emph{Y}, and confounding factor \emph{C}. The direct link denotes the causalities between the two nodes.

In Figure 1(b), it exists a backdoor path, \emph{T} $\leftarrow$ \emph{C} $\rightarrow$ \emph{Y}. This means that the previous methods may mistakenly learn the spurious correlation brought by the backdoor path \emph{T} $\leftarrow$ \emph{C} $\rightarrow$ \emph{Y}. In this paper, we propose the causal interventions-based model. We used the context-based intervention to block spurious correlation between contexts and class labels. In the causal inference view, as Figure 1(c) shows, we can use $P(E\big|do(T))$ instead of $P(E\big|T)$ to avoid the effect of the confounding factors. The way $do(T)$ (\emph{do}-operation) can pursue the causality between the contexts \emph{T} and the class labels \emph{Y} without the confounder \cite{pearl2009causality}. This $P(Y\big|E,do(T))$ blocks the backdoor path and eliminates the spurious correlation from the confounding factor. We may intervene \emph{T} to calculate $P(Y\big|E,do(T))$ by randomised controlled trial \cite{chalmers1981method}. It needs to be pointed out that it is difficult to obtain the sample-control on the contexts for the one-shot task. Inspired by incremental learning \cite{kohn2018incrementa,rebuffi2017icar,thrun1995learning,french1999catastrophic,de2019episodic}, we use the previous knowledge as the previous prototype to intervene in the previous and current prototypes as the final prototype for the one-shot task. This way also avoids catastrophic forgetting \cite{thrun1995learning} of the model. Our contributions can be summarized as follows:

\begin{itemize}
\item We propose the causal model from a causal perspective, intervening on the contexts, $do(T)$ (\emph{do}-operation) contexts and prototypes. This way helps the model prevent overfitting to the current data. Our method eliminates spurious correlations brought by the few sample selection bias and enhances the model's generalization.

\item To better distinguish predefined entity classes and other class, we perform preliminary identification of entities through span, which helps the model to identify the boundaries of entities better and helps in the final type identification. Different from only considering the features within classes, our span considers the features among classes.

\item We evaluated on Few\_NERD and achieved state-of-the-art results. Comprehensive experiments show that our model has a more robust generalization. Our approach achieves 
up to 29\% absolute improvement and 12\% on average for all tasks.
\end{itemize}

\section{Related Work}
\label{intro}

\subsection{Few-Shot Learning and Meta-Learning}

Recently, few-shot learning has been widely used in natural language processing \cite{chen2019meta,gao2020making,brown2020language,schick2020s}. The obvious problem of few-shot learning is overfitting, so researchers usually introduce source domain data \cite{han2018fewrel,geng2019induction,wang2021behind}. Besides, few-shot learning usually introduces meta-learning to deal with it. Meta-learning was first widely used in computer vision. With the proposal of prototype networks, metric-based \cite{kulis2013metric,vinyals2016matching,snell2017prototypical} methods were widely adopted. The method first encodes vectors in the support set, after getting the prototype representation of each class based on all vectors of the same class. When predicting query, it calculates the distance between the prototype and query data using different metrics (generally Euclidean distance), and finally classifies the query data according to the nearest prototype. Prototype networks have been well performed in many tasks of natural language processing.

\subsection{Few-shot Named Entity Recognition}

Few-shot Named Entity Recognition aims to use a few support samples to make the model able to recognize unknown entities in the training phase. Previous work has seen many approaches to token-level  \cite{fritzler2019few,yang2020simple,hou2020few}. \citet{snell2017prototypical} used a prototype network for few-shot Named Entity Recognition. Later, inspired by feature extraction and nearest neighbors, \citet{yang2020simple} propose NNShot and StructShot. NNShot uses the nearest neighbor to classify entities. Based on this, StructShot added viterbi to improve the few-shot Named Entity Recognition task. \citet{ding2021few} presented Few-NERD, a large-scale human-annotated few-shot NER dataset. And it was evaluated in ProtoBERT, NNShot and StructShot methods. In addition, prompt-based \cite{cui2021template} technologies are also appearing in this area. And \citet{tong2021learning} proposes multiple prototypes for reasoning. However, these methods only learn the semantic features and intermediate representations of classes from the source domain. But the generalization to the target domain is very low. So \citet{das2021container} proposes to insert Gaussian embedding and contrast learning to improve the accuracy of few-shot Named Entity Recognition. These works mark and assign labels to entities, but ignore the integrity and boundedness of entities. Therefore, span-level \cite{wang2021enhanced,yu2021few,athiwaratkun2020augmented,wang2021learning} approaches has been proposed. ESD \cite{wang2021enhanced} proposes to use span representation and span matching to enhance the completeness of entity recognition. Based on entity span, \citet{ma2022decomposed} argues that the previous approach contains O-class noise and lacks parameter updates during transfer. Therefore, \citet{ma2022decomposed} proposes to fine-tune the parameters using support instances and locate only entities during span detection.

However, all these methods consider only the current support and query instances, causing the model forgets the previous data. Also, the model ignores the impact of context on entities, which can lead to the overfitting of contexts and entities. And for support and query instances, the model only makes a simple metric calculation, without considering both jointly. We believe that different instances in support contribute differently to query and cannot be calculated with equal weights.

\subsection{Causal Inference}

The purpose of causal inference \cite{pearl2009causality} is to remove the confounders between variables and get causal effects. Based on the causal effects between variables, make accurate predictions of the task. Considering the fitting problem and causal effects of the few samples, we aim to use causal inference to improve the robustness and transferability of the model. Many works have been done to improve the robustness \cite{zeng2020counterfactual,tang2020unbiased,wang2020visual,qi2020two} of models using causal theory. Also, in causal effects, many studies use front door adjustment or back door adjustment to remove spurious correlations from confounders \cite{yue2020interventional,tang2020long,zhang2020causal,zhang2021biasing,liu2021element} and find the causal effects of variables.

\begin{figure*}[h]
  \centering
  \includegraphics[width=0.9\textwidth,height=0.32\textwidth]{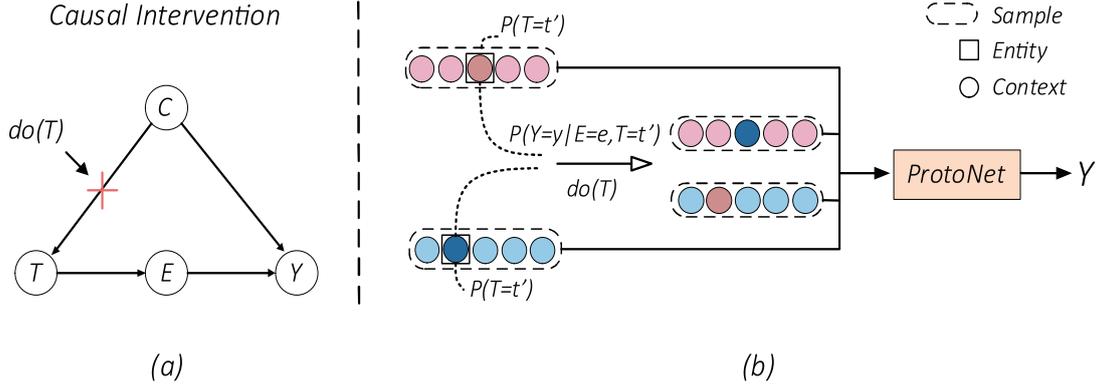}
  \caption{(a) The causal structure, where \emph{T} is the context, \emph{E} is the entity representation, \emph{Y} is the predicted label. Besides, there exists \emph{C}, the confounders, such as samples selection bias. (b) Our context-based intervention, to block the spurious correlation between context and predicted label. We replaces entities of the same type to get a new sample, and feeds into the prototype network to get the predicted label.}\label{1}
\end{figure*}


\section{Task Formulation}

NER is usually formulated as a sequence labeling problem. For a sequence \emph{\{x$_{1}$.. .x$_{n}$\}}, NER aims to assign a label for each \emph{x$_{i}$} . The labels indicate whether \emph{x$_{i}$} belongs to a named entity (such as person, location) or does not belong to any entity (O class).

Few NER uses a few data to identify unknown classes in training by taking N-way K-shot. In which, Few NER sets N-way K-shot by iteratively constructing episodes. At the sampling of training, each episode includes \emph{N} classes and each class contains \emph{K} samples, thus constructing the support set, \emph{S$_{train}$ = \{(x$^{(i)}$,y$^{(i)}$)\}$^{N*Q}_{i=1}$}. For the construction of query set, \emph{Q} samples of \emph{N} classes are sampled, \emph{Q$_{train}$ = \{ (x$^{(i)}$,y$^{(i)}$)\}$^{N*Q}_{i=1}$}. Where \emph{S$_{train}$ $\bigcap$ Q$_{train}$ =$\varnothing$}.

According to Strain prediction \emph{Q$_{train}$}, which makes the model get trained. For testing, we use a few \emph{S$_{test}$ } and make predictions for \emph{Q$_{test}$}. Similarly, \emph{S$_{test}$ $\bigcap$ Q$_{test}$=$\varnothing$}. Note that the class of entities in the test set is not present in the training set, that is, \emph{Y$_{train}$ $\bigcap$ Y$_{test}$=$\varnothing$}.

\section{Methods}
\label{intro}

In this part, we show our method for the few named entity recognition. The method is composed of four main parts: entity span detection, context-based intervention, prototype-based intervention, and entity reweighting.

\subsection{Causal Intervention}

In the few-shot named entity recognition task, the serious selection bias of data causes spurious correlation, and then misleads the overfitting of the model. We use the causal interventions-based method to solve the spurious correlation. Figure 2 illustrates the problem in the task. \emph{T} represents the context, \emph{E} represents the entity , \emph{Y} is the predicted label, and \emph{C} is the confounders, such as the few samples selection bias.

\textbf{\emph{T}$\rightarrow$ \emph{E}$\rightarrow$ \emph{Y}}   
The entity representation is learned in terms of contexts. Model uses the entity representation to get the predicted label. 

\textbf{\emph{T}$\leftarrow$ \emph{C}$\rightarrow$ \emph{Y}}  A backdoor path that leads to a spurious correlation in the model. \emph{C} represents all possible confounders in the task, such as the few samples selection bias. These bias make context pay more attention to the current data and mislead prediction labels. This backdoor path caused the spurious correlation between the context and the predicted class label. The spurious correlation further leads to the overfitting of context and predicted label.

Then, considering that in 1-shot, no additional entities can intervene, we analyze the model. Figure 3 illustrates the problem in the model. In the two figures, \emph{T} represents the context, \emph{E} represents the entity , \emph{P} is the entity prototype, \emph{Y} is the predicted label, and \emph{C} is the confounders.

\textbf{\emph{P}$\rightarrow$ \emph{Y}}  The model calculates the Euclidean distance through the entity prototype, and gets the the predicted label.

\textbf{\emph{P}$\leftarrow$ \emph{C}$\rightarrow$ \emph{Y}} Confounders such as samples selection bias, mislead the calculation of entity prototype, resulting in the spurious correlation between prototype and predicted label. This spurious correlation leads to overfitting.

We expect to block the backdoor path \emph{T}$\leftarrow$ \emph{C}$\rightarrow$ \emph{Y} and \emph{P}$\leftarrow$ \emph{C}$\rightarrow$ \emph{Y}. So, we intervene on \emph{T} and \emph{P}. Considering \emph{C} as the confounders are hard to catch. Therefore, we use the front-door adjustment for the calculation, shown in Eq 1. We get the final equation 2. Specific derivation details are in the Appendix. Here, in \emph{$\textstyle\sum_{E}P(E=e|T=t)$}, \emph{E} is the entity representation. If we consider \emph{E} as a binary classification, entity and non-entity, then this part denotes the entity detection. In \emph{$\sum_{T'}P(Y=y|E=e,T=t')$}, \emph{$T=t'$} means we need to iterate over each \emph{T} to get the predicted label \emph{Y} in the case of \emph{E=e}. Therefore, we propose a context-based intervention, make the entity replacement. In 1-shot, no additional entity to intervene, so we intervene on the prototype, consider the previous and current prototypes. Besides, in \emph{$P(T=t')$}, it denotes the result of selecting the current \emph{T}. We calculate the weights to get the prototype, considering that different \emph{T} may be different from the original \emph{T} distribution.  Finally, we divide the model into four parts: entity detection, context-based causal intervention, prototype-based causal intervention and sample reweighting.

\begin{small}
\begin{align}
    P(Y=y|do(T&=t))=\sum_{E}P(E=e|do(T=t))
    \notag
    \\&P(Y=y|do(T=t),E=e)
\end{align}
\end{small}

\begin{small}
\begin{align}
    P(Y&=y|do(T=t))=\sum_{E}P(E=e|T=t)
    \notag
    \\ &\sum_{t'}P(Y=y|E=e,T=t')P(T=t')
\end{align}
\end{small}

\begin{figure*}
  \centering
  \includegraphics[width=0.95\textwidth]{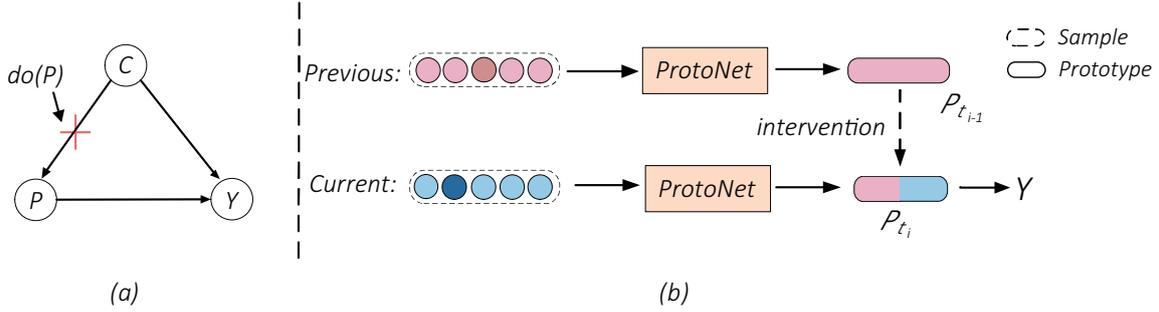}
  \caption{(a) Prototype-based intervention, where \emph{P} is the entity prototype, \emph{Y} is the predicted label, \emph{C} is the confounders. (b) We intervene the prototype to block the spurious correlation between the prototype and predicted label. During the current episode, the previous prototype is introduced to the current episode as knowledge. The two prototypes are combined as the current prototype. In the next episode, this prototype is introduced as the previous prototype.}\label{1}
\end{figure*}

\subsection{Entity Detection}

Due to the \emph{$\sum_{E}P(E=e|T=t)$}, we propose the entity detection. The traditional prototype network is to average all vectors of the same class entities after encoding, as a prototype representation of each class. The typical prototype network considers only the features common in each class, but not the commonality cross classes. The span detection aims to locate all the named entities in the input sequence but without differentiating specific classes. It considers the possible similarity between entities of all classes. We feed all the entities as a whole into the prototype network, and likewise all the non-entities, to determine all the entities in a sequence. In this way, we expect to discover features common to all entities to help us make a simple initial filtering of the input. An important note is that in the predicted label integration, the dimension matching problem needs to be considered. Therefore, we expand the entity judgment results to the same dimension as the class, to achieve dimension matching.

\subsection{Context-based Causal Intervention}

The few-shot task has limited samples in each episode, so the model will easily overfit the current samples. Such overfitting is misleading due to spurious correlation caused by samples selection bias. Within the sample, the entities are included in the contexts. When the sample is limited, the representation of entities can easily overfit the current context, making the model difficult to generalize to new domains.

Figure 2 shows our method. In Fig 2(a), we define the few-shot named entity recognition task. In this task, the confounders, such as the few samples selection bias, cause a backdoor path \emph{T}$\leftarrow$ \emph{C}$\rightarrow$ \emph{Y} between context and predicted label. This backdoor path caused the spurious correlation between context and predicted label, misleading context and label overfitting. We want to block this backdoor path. Therefore, we intervene on T to block the backdoor path \emph{T}$\leftarrow$ \emph{C}$\rightarrow$ \emph{Y}. We refer to this method as the context-based causal intervention.

The method is gotten according to Eq 2. In \emph{$\sum_{t'}P(Y=y|E=e,T=t')$}, the \emph{$T=t'$} means we need to iterate over each \emph{T} to get the final \emph{Y} in case \emph{E=e}. In other words, under each entity, we have to iterate over all \emph{T} of the same type. Therefore, it is shown in Fig 2(b). In each sentence, the entities in the sentence are replaced with other entities of the same type in turn for the purpose of traversal. This process we use only in training, in testing we use the original data.

\subsection{Prototype-based Causal Intervention}

When performing 1-shot experiment, context-based causal intervention is not possible since there is no additional suitable entity to intervene. Here, we consider another way, prototype-based intervention. Figure 3 shows the method.

The previous methods average the current episode support as the class prototype representation, and use the distance between query and prototype to recognize the entities. However, as Figure 3(a) shows, there exist confounders, such as the few samples selection bias. It leads to the backdoor path between the prototype and the label, causing the spurious correlation. This spurious correlation results in overfitting in the calculation of entity prototypes. So we refine the model process from a causal perspective. Our method blocks the path \emph{P}$\leftarrow$ \emph{C}$\rightarrow$ \emph{Y} by intervening \emph{P}, a method called prototype-based causal intervention. Specifically, as shown in Fig 3(b), we save the prior prototype as knowledge. In the current episode, the previous knowledge is combined with the current class representation as the current class representation. Then the same calculation is repeated in the next episode.

Through the prototype-based intervention, we block the spurious correlation between the prototype and the predicted label. This method considers both current and previous data. Also, it prevents the model from overfitting the entity prototype calculation.

\subsection{Sample Reweighting}

Due to \emph{$P(T=t')$}, we propose sample reweighting. Typical meta-learning for prototype calculation defaults the weights of all samples to 1. We argue that each sample provides a different contribution to the prototype calculation. 

Specifically, for each support sample, we calculate the distance between it and the query. After that, using softmax, we transform a weight for each sample, as shown in Eq 3. Finally, the weights are applied to each sample to calculate the entity prototype. Meanwhile, we also add Maximum mean discrepancy(MMD) to the loss calculation, taking the training data as the source domain and the test support data as the target domain, and calculate the difference between the source and target domains by MMD. By double calculation, we reduce the distribution difference between the source and target domains. Regarding the loss function, the MMD is used as the final feedback loss with the classification loss, as shown in Eq 4.

\begin{align}
    \alpha_{i}=softmax(h_{\theta}(x_{q})-h_{\theta}(x_{s_{i}}))
\end{align}
\begin{align}
    L(\theta)=\frac{1}{N}\sum_{i=1}^{N}CrossEntropy(y_{i},h_{\theta}(x_{i}))
    \notag
    \\ +\sup_{\|f\|_{H}\leq 1}E_{p}[f(s)]-E_{q}[f(t)]
\end{align}

Where \emph{h} is our network, \emph{y} is the true label, \emph{p} denotes the distribution of the source domain \emph{s}, and \emph{q} denotes the distribution of the target domain \emph{t}.

\section{Experiments}
\label{intro}

\begin{table*}
\tiny
  \centering
  \begin{spacing}{1.2}
    \setlength{\tabcolsep}{0.8mm}{
    \begin{tabular}{ccccccccccccc}
    \toprule
    \multicolumn{13}{c}{\textbf{Few-NERD(INTRA)}} \\
    \midrule
    \multicolumn{1}{c}{\multirow{2}{*} {\textbf{Model}}}
    &\multicolumn{3}{c}{\textbf{5-way 1$ \sim $2-shot}}
    &\multicolumn{3}{c}{\textbf{5-way 5$ \sim $10-shot}}
    &\multicolumn{3}{c}{\textbf{10-way 1$ \sim $2-shot}}
    &\multicolumn{3}{c}{\textbf{10-way 5$ \sim $10-shot}}
    \\
    \cmidrule(lr){2-4}     \cmidrule(lr){5-7}      \cmidrule(lr){8-10}     \cmidrule(lr){11-13}
    \multicolumn{1}{c}{}
    &\multicolumn{1}{c}{P}&{R}&{F1}
    &\multicolumn{1}{c}{P}&{R}&{F1}
    &\multicolumn{1}{c}{P}&{R}&{F1}
    &\multicolumn{1}{c}{P}&{R}&{F1}
    \\
    \cmidrule(lr){1-1}      \cmidrule(lr){2-4}     \cmidrule(lr){5-7}      \cmidrule(lr){8-10}     \cmidrule(lr){11-13}
    ProtoBERT &16.35$\pm$0.63 &28.35$\pm$2.41 &20.71$\pm$1.16 &31.43$\pm$1.14 &45.28$\pm$0.71 &37.08$\pm$1.01 &12.05$\pm$1.09 &21.27$\pm$1.35 &15.32$\pm$0.68 &23.15$\pm$0.42 &35.83$\pm$0.97 &28.02$\pm$0.56 \\
    NNShot &20.47$\pm$0.40 &23.05$\pm$1.12 &21.58$\pm$0.70 &23.88$\pm$0.79 &28.35$\pm$0.88 &25.66$\pm$0.78 &14.83$\pm$0.56 &16.90$\pm$0.68 &15.72$\pm$0.53 &18.18$\pm$1.20 &22.45$\pm$1.03 &19.82$\pm$1.11 \\
    StructShot &31.40$\pm$1.34 &19.63$\pm$2.61 &23.95$\pm$2.39 &45.20$\pm$1.08 &22.80$\pm$0.99 &29.68$\pm$1.11 &23.15$\pm$0.77 &8.61$\pm$0.69 &12.31$\pm$0.72 &40.40$\pm$2.46 &11.35$\pm$1.32 &17.10$\pm$1.75 \\
    CONTAINER &38.46$\pm$0.55 &41.66$\pm$0.23 &40.00$\pm$0.71 &47.50$\pm$0.61 &63.33$\pm$0.24 &54.28$\pm$0.64 &38.33$\pm$0.45 &33.33$\pm$0.27 &35.89$\pm$0.16 &43.95$\pm$0.22 &54.05$\pm$0.73 &48.48$\pm$0.69 \\
    ESD &42.94$\pm$4.47 &32.69$\pm$1.55 &37.12$\pm$0.89 &59.55$\pm$0.89 &43.83$\pm$3.08 &50.50$\pm$1.79 &37.58$\pm$1.53 &26.76$\pm$1.96 &31.26$\pm$0.53 &34.89$\pm$2.75 &32.00$\pm$3.42 &33.38$\pm$4.71 \\
    DML &47.30$\pm$1.94 &46.64$\pm$1.27 &46.97$\pm$2.36 &57.70$\pm$2.48 &61.20$\pm$1.33 &59.40$\pm$2.48 &40.36$\pm$0.67 &42.30$\pm$0.73 &41.30$\pm$0.89 &51.69$\pm$2.48 &49.77$\pm$1.64 &50.71$\pm$4.24 \\
    \textbf{Ours} &\textbf{52.44$\pm$1.01} &\textbf{55.67$\pm$1.29} &\textbf{54.00$\pm$1.14} &\textbf{85.06$\pm$1.31} &\textbf{88.92$\pm$1.09} &\textbf{86.94$\pm$1.21} &\textbf{42.23$\pm$0.52} &\textbf{43.81$\pm$0.73} &\textbf{43.58$\pm$1.20} &\textbf{75.92$\pm$1.16} &\textbf{82.34$\pm$0.73} &\textbf{79.00$\pm$0.96} \\
    \bottomrule
    \end{tabular}%
    }
    \end{spacing}
  \caption{Performance of state-of-art models on Few-NERD (INTRA)}
  \label{tab:addlabel}%
\end{table*}%

\begin{table*}
\tiny
  \centering
  \begin{spacing}{1.2}
    \setlength{\tabcolsep}{0.8mm}{
    \begin{tabular}{ccccccccccccc}
    \toprule
    \multicolumn{13}{c}{\textbf{Few-NERD(INTER)}} \\
    \midrule
    \multicolumn{1}{c}{\multirow{2}{*} {\textbf{Model}}}
    &\multicolumn{3}{c}{\textbf{5-way 1$ \sim $2-shot}}
    &\multicolumn{3}{c}{\textbf{5-way 5$ \sim $10-shot}}
    &\multicolumn{3}{c}{\textbf{10-way 1$ \sim $2-shot}}
    &\multicolumn{3}{c}{\textbf{10-way 5$ \sim $10-shot}}
    \\
    
    \cmidrule(lr){2-4}     \cmidrule(lr){5-7}      \cmidrule(lr){8-10}     \cmidrule(lr){11-13}
    \multicolumn{1}{c}{}
    &\multicolumn{1}{c}{P}&{R}&{F1}
    &\multicolumn{1}{c}{P}&{R}&{F1}
    &\multicolumn{1}{c}{P}&{R}&{F1}
    &\multicolumn{1}{c}{P}&{R}&{F1}
    \\
    \cmidrule(lr){1-1}      \cmidrule(lr){2-4}     \cmidrule(lr){5-7}      \cmidrule(lr){8-10}     \cmidrule(lr){11-13}
    ProtoBERT &31.45$\pm$0.74 &46.44$\pm$3.40 &37.49$\pm$1.63 &46.88$\pm$0.27 &59.54$\pm$1.10 &52.42$\pm$0.60 &22.17$\pm$0.92 &34.72$\pm$0.52 &26.98$\pm$0.79 &50.87$\pm$1.01 &63.30$\pm$0.66 &56.29$\pm$0.79 \\
    NNShot &38.32$\pm$2.24 &42.82$\pm$2.34 &40.31$\pm$2.30 &39.40$\pm$1.42 &43.34$\pm$7.32 &42.66$\pm$1.07 &29.52$\pm$1.15 &34.06$\pm$2.27 &31.54$\pm$1.63 &33.74$\pm$0.44 &41.82$\pm$0.52 &37.09$\pm$0.13 \\
    StructShot &49.45$\pm$0.60 &32.44$\pm$7.77 &38.78$\pm$5.70 &42.62$\pm$6.46 &32.47$\pm$5.37 &35.95$\pm$1.09 &32.54$\pm$1.42 &17.54$\pm$0.72 &22.61$\pm$0.95 &41.82$\pm$0.50 &44.52$\pm$0.74 &42.75$\pm$0.62 \\
    CONTAINER &46.15$\pm$2.47 &60.00$\pm$1.98 &52.17$\pm$2.74 &56.41$\pm$0.87 &66.66$\pm$0.53 &61.11$\pm$0.76 &43.75$\pm$1.16 &58.33$\pm$1.27 &50.00$\pm$1.46 &57.57$\pm$0.46 &62.29$\pm$0.75 &59.84$\pm$0.36 \\
    ESD &61.39$\pm$2.83 &54.17$\pm$2.07 &57.56$\pm$2.52 &76.70$\pm$3.63 &64.24$\pm$1.77 &69.92$\pm$0.56 &58.37$\pm$5.01 &48.35$\pm$1.94 &52.89$\pm$1.11 &68.72$\pm$0.14 &65.18$\pm$0.35 &66.90$\pm$0.60 \\
    DML &62.76$\pm$2.47 &62.76$\pm$2.08 &62.76$\pm$2.61 &69.87$\pm$0.45 &71.96$\pm$1.43 &70.90$\pm$0.28 &57.54$\pm$1.87 &61.74$\pm$2.04 &59.57$\pm$2.88 &63.15$\pm$0.33 &70.81$\pm$1.01 &66.76$\pm$0.54 \\
    \textbf{Ours} &\textbf{65.96$\pm$0.36} &\textbf{73.23$\pm$2.29} &\textbf{69.41$\pm$1.24} &\textbf{78.74$\pm$0.38} &\textbf{85.29$\pm$0.28} &\textbf{81.89$\pm$0.33} &\textbf{59.26$\pm$0.90} &\textbf{62.14$\pm$0.52} &\textbf{60.67$\pm$0.73} &\textbf{82.13$\pm$0.15} &\textbf{83.58$\pm$0.24} &\textbf{82.13$\pm$0.10} \\
    \bottomrule
    \end{tabular}%
    }
    \end{spacing}
  \caption{Performance of state-of-art models on Few-NERD (INTER)}
  \label{tab:addlabel}%
\end{table*}%

\subsection{Dataset}

Few-NERD\cite{ding2021few} It includes an annotation structure with 8 coarse-grained entity types and 66 fine-grained entity types. Based on this, two tasks are designed: i) FewNERD-INTRA, where all entities in the training set (source domain), validation set and test set (target domain) belong to different coarse-grained types. ii) FewNERD-INTER, where the training set, validation set and test set can share coarse-grained types, but the fine-grained entity types are disjoint.
FewNERD uses \emph{N-way K$ \sim $2K shots}. FewNERD-INTRA and FewNERD-INTER both have four settings. \emph{5-way 1$ \sim $2-shot}, \emph{5-way 5$ \sim $10-shot}, \emph{10-way 1$ \sim $2-shot} and \emph{10-way 5$ \sim $10-shot}.

\subsection{Parameter Settings}
Following previous methods\cite{ding2021few}, we used the Bert-base-uncased model \cite{devlin2018bert}. We used maximum sequence length of 32, and set the dropout ratio to 0.1. We used AdamW \cite{loshchilov2017decoupled}. More details of the parameter settings, please refer to the Appendix.

\subsection{Evaluation metrics}

For the evaluation of Few-NERD, we followed \cite{ding2021few}, calculated the pecision(\emph{P}), recall(\emph{R}) and micro F1-score(\emph{F1}).

\subsection{Baselines}

For systematic comparison, we have chosen a variety of methods, including: ProtoBERT, NNShot, StructShot, CONTAINER\cite{das2021container}, ESD\cite{wang2021enhanced} and Decomposed Meta-Learning\cite{ma2022decomposed}. Please refer to the Appendix for baselines specific details.

\subsection{Results and Analysis}
The table shows the results of the model in the FewNERD-INTRA and FewNERD-INTER datasets of Few-NERD, respectively. Comprehensive experiments show that our method achieves state-of-the-art. By comparing the results, our method improves by 11-29\% on 5-shot, which demonstrates the effectiveness of context-based causal intervention and entity detection. The method also improved by 1-8\% on 1-shot, demonstrating the advantages of prototype-based causal intervention and sample reweighting. This shows that our method can block the spurious correlation in the task and prevent overfitting.


\subsection{Ablation Study}
In order to evaluate the contribution of the different components of the proposed method, we performed the following baseline as an ablation study: for 1$ \sim $2-shot, 1) We use the basic prototype network and do the entity detection, prototype-based intervention and sample reweighting. 2) Without prototype-based intervention, we use the basic prototype network and do the entity detection and sample reweighting. 3) Without entity detection, we only use the basic prototype network and do sample reweighting and prototype-based intervention. 4) We only use the basic prototype network and do the prototype-based intervention. 5) We only do the sample reweighting. For 5$ \sim $10-shot, 1) Without sample reweighting, we make a context-based intervention and do the entity detection. 2) We only make a context-based intervention for the basic prototype network. 3) We only use the basic prototype network and do the entity detection.

The table illustrates the contribution of each component in our proposed method. The effect decreases when any of the components are removed. Also, by observation, we have the following findings. Our entity detection better corrects and improves entity recognition, making a 6-19\% improvement. Meanwhile, context-based causal intervention can help us better resolve spurious correlation, which can improve 20\%. For 1-shot, our prototype-based causal intervention can also play a great role, an increase of 6\%. In addition, it shows that sample reweighting can help the model to improve by 18\%.

\begin{table}[h]
\tiny
  \centering
  \begin{spacing}{1.0}
  \setlength{\tabcolsep}{0.85mm}{
    \begin{tabular}{lccccc}
    \toprule
    \multicolumn{1}{c}{\multirow{2}{*}{\textbf{Setting}}}
    &\multicolumn{1}{c}{\textbf{Entity}}
    &\multicolumn{1}{c}{\textbf{Context-based}}
    &\multicolumn{1}{c}{\textbf{Sample}}
    &\multicolumn{1}{c}{\textbf{Prototype-based}}
    &\multicolumn{1}{c}{\textbf{F1}}
    \\
    \multicolumn{1}{l}{}
    &\multicolumn{1}{c}{\textbf{Detection}}
    &\multicolumn{1}{c}{\textbf{Intervention}}
    &\multicolumn{1}{c}{\textbf{Reweighting}}
    &\multicolumn{1}{c}{\textbf{Intervention}}
    &\multicolumn{1}{c}{\textbf{score}}
    \\
    \cmidrule(lr){1-6}
    \multirow{5}{*} {1$ \sim $2-shot} 
    &\checkmark & $ \times $ & \checkmark & \checkmark & 69.41 \\
    \multicolumn{1}{l}{} &\checkmark &$\times$ &\checkmark &$\times$ &63.22\\
    \multicolumn{1}{l}{} &$\times$ &$\times$ &\checkmark &\checkmark & 60.54\\
    \multicolumn{1}{l}{} &$\times$ &$\times$ &$\times$ &\checkmark & 58.26\\
    \multicolumn{1}{l}{} &$\times$ &$\times$ &\checkmark &$\times$ & 57.64\\
    \cmidrule(lr){1-6}
    \multirow{3}{*} {5$ \sim $10-shot} 
    &\checkmark &\checkmark &$\times$ &$\times$ & 81.89 \\
    \multicolumn{1}{l}{} &$\times$ &\checkmark &$\times$ &$\times$ &75.40\\
    \multicolumn{1}{l}{} &\checkmark &$\times$ &$\times$ &$\times$ & 61.67\\
    \bottomrule
    \end{tabular}%
    }
    \end{spacing}
  \caption{Ablation study: F1 scores on Few-NERD}
  \label{tab:addlabel}%
\end{table}%

\subsection{Experimental Analysis}

\textbf{How does entity detection improve entity recognition} To further illustrate the role of entity detection, an example is given. Entity detection performs well on some boundary information classifications, and it helps identify entities and correct misidentifications. For example with the identification of \emph{'the'}, entity detection can help to identify \emph{'the'} in \emph{'The Porcellian Club'} and \emph{'The Nation Game'}, while ignoring \emph{'the year'} and \emph{'the club'}. Similarly, for \emph{American}, entity detection can help us identify the League after \emph{American} in \emph{'American League'} while ignoring the airline after \emph{American} in \emph{'American airline'}.

\begin{figure}[h]
  \centering
  \includegraphics[width=0.4\textwidth]{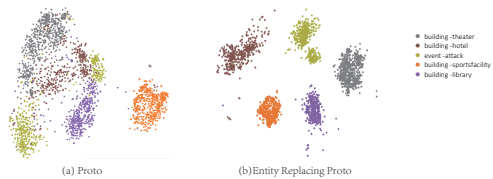}
  \caption{t-SNE visualization of entity replacing on Few-NERD Intra, 5-way 5$ \sim $10-shot.}\label{1}
\end{figure}

\textbf{How does context-based causal intervention improve entity recognition} To alleviate the overfitting problem of the model, we propose entity replacement. To further demonstrate that context-based causal intervention solves the overfitting problem, as shown in figure 4, we give the t-SNE of the original prototype network and the prototype network with the entity replaced. In the original prototype network, the embedding is very scattered, and the boundary is not obvious, it is easy to confuse. As we do the context-based causal intervention, the embedding becomes more compact and there are clear boundaries. It makes entities of the same type embedded closer together and alienates the distance between different types. The optimization of embedded space enhances the ability of the few-shot named entity recognition.


\textbf{How does prototype-based causal intervention improve entity recognition} For example, in the current query, when we want to judge the class of \emph{German} in \emph{'The local dialect of the region is East Franconian German, referred to in German as Frankisch'}. Since \emph{German} is O-class (other class) in the current support, the machine will classify German as O-class. This is an overfitting of the current prototype calculation. When we intervene in the prototype, the effect will improve. For example, in the previous support data, there was a case where \emph{German} was classified as other-lanauage. We transformed it into a previous prototype, and introduced it in the current episode. By prototype-based causal intervention, the machine can correct this error and correctly classify \emph{German} as other-lanauage in the current query.

\begin{figure}[h]
  \centering
  \includegraphics[width=0.4\textwidth]{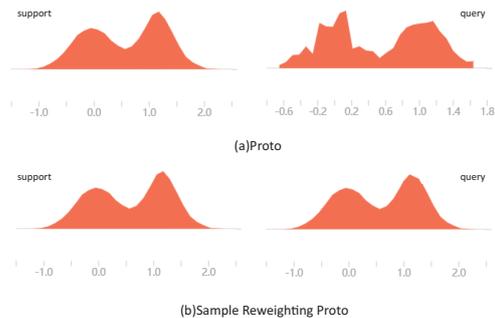}
  \caption{histogram of sample reweighting on Few-NERD Intra, 5-way 5$ \sim $10-shot. }\label{1}
\end{figure}

\textbf{How does sample reweighting improve entity recognition} As shown in figure 5, we show the data distribution histograms of the source domain and the target domain in the original prototype network and after sample reweighting. Through the comparison, after sample reweighting, the distributions of the source domain and the target domain are basically consistent, and the gap is reduced. This improvement enables the model to transfer from the source domain to the target domain well, and improves the effect of few-shot named entity recognition.

\section{Conclusion and Future Work}
In this paper, we analyze the few-shot named entity recognition from a causal perspective. There exists the spurious correlation between the context and predicted label, which causes the overfitting between context and predicted label. Meanwhile, there exists a spurious correlation between the prototype and the predicted label, which causes the overfitting between prototype calculation and the predicted label. Therefore, we propose a causal interventions-based few-shot named entity recognition method. Firstly, the boundary of the entity is detected. Then, make a context-based causal intervention, and replaces the entity to add a new context. For 1-shot, the context-based causal intervention cannot be applicable, we further propose the prototype-based causal intervention, considering previous and current prototypes. Finally, we use the support and query to calculate weights to reweight the samples. Comprehensive experiments show that our method achieves significant improvement and helps models avoid overfitting and transfer to new domains better.

In addition, the method still has some weaknesses. The time consumption of the method is small, but the memory consumption is large. In the future, we will continue to improve the method and reduce memory consumption.


\bibliography{anthology}

\begin{thebibliography}{42}
\expandafter\ifx\csname natexlab\endcsname\relax\def\natexlab#1{#1}\fi

\bibitem[{Athiwaratkun et~al.(2020)Athiwaratkun, Santos, Krone, and
  Xiang}]{athiwaratkun2020augmented}
Ben Athiwaratkun, Cicero Nogueira~dos Santos, Jason Krone, and Bing Xiang.
  2020.
\newblock Augmented natural language for generative sequence labeling.
\newblock \emph{arXiv preprint arXiv:2009.13272}.

\bibitem[{Brown et~al.(2020)Brown, Mann, Ryder, Subbiah, Kaplan, Dhariwal,
  Neelakantan, Shyam, Sastry, Askell et~al.}]{brown2020language}
Tom Brown, Benjamin Mann, Nick Ryder, Melanie Subbiah, Jared~D Kaplan, Prafulla
  Dhariwal, Arvind Neelakantan, Pranav Shyam, Girish Sastry, Amanda Askell,
  et~al. 2020.
\newblock Language models are few-shot learners.
\newblock \emph{Advances in neural information processing systems},
  33:1877--1901.

\bibitem[{Chalmers et~al.(1981)Chalmers, Smith~Jr, Blackburn, Silverman,
  Schroeder, Reitman, and Ambroz}]{chalmers1981method}
Thomas~C Chalmers, Harry Smith~Jr, Bradley Blackburn, Bernard Silverman, Biruta
  Schroeder, Dinah Reitman, and Alexander Ambroz. 1981.
\newblock A method for assessing the quality of a randomized control trial.
\newblock \emph{Controlled clinical trials}, 2(1):31--49.

\bibitem[{Chen et~al.(2019)Chen, Zhang, Zhang, Chen, and Chen}]{chen2019meta}
Mingyang Chen, Wen Zhang, Wei Zhang, Qiang Chen, and Huajun Chen. 2019.
\newblock Meta relational learning for few-shot link prediction in knowledge
  graphs.
\newblock \emph{arXiv preprint arXiv:1909.01515}.

\bibitem[{Chiu and Nichols(2016)}]{chiu2016named}
Jason~PC Chiu and Eric Nichols. 2016.
\newblock Named entity recognition with bidirectional lstm-cnns.
\newblock \emph{Transactions of the association for computational linguistics},
  4:357--370.

\bibitem[{Cui et~al.(2021)Cui, Wu, Liu, Yang, and Zhang}]{cui2021template}
Leyang Cui, Yu~Wu, Jian Liu, Sen Yang, and Yue Zhang. 2021.
\newblock Template-based named entity recognition using bart.
\newblock \emph{arXiv preprint arXiv:2106.01760}.

\bibitem[{Das et~al.(2021)Das, Katiyar, Passonneau, and
  Zhang}]{das2021container}
Sarkar Snigdha~Sarathi Das, Arzoo Katiyar, Rebecca~J Passonneau, and Rui Zhang.
  2021.
\newblock Container: Few-shot named entity recognition via contrastive
  learning.
\newblock \emph{arXiv preprint arXiv:2109.07589}.

\bibitem[{de~Masson~D'Autume et~al.(2019)de~Masson~D'Autume, Ruder, Kong, and
  Yogatama}]{de2019episodic}
Cyprien de~Masson~D'Autume, Sebastian Ruder, Lingpeng Kong, and Dani Yogatama.
  2019.
\newblock Episodic memory in lifelong language learning.
\newblock \emph{Advances in Neural Information Processing Systems}, 32.

\bibitem[{Devlin et~al.(2018)Devlin, Chang, Lee, and
  Toutanova}]{devlin2018bert}
Jacob Devlin, Ming-Wei Chang, Kenton Lee, and Kristina Toutanova. 2018.
\newblock Bert: Pre-training of deep bidirectional transformers for language
  understanding.
\newblock \emph{arXiv preprint arXiv:1810.04805}.

\bibitem[{Ding et~al.(2021)Ding, Xu, Chen, Wang, Han, Xie, Zheng, and
  Liu}]{ding2021few}
Ning Ding, Guangwei Xu, Yulin Chen, Xiaobin Wang, Xu~Han, Pengjun Xie, Hai-Tao
  Zheng, and Zhiyuan Liu. 2021.
\newblock Few-nerd: A few-shot named entity recognition dataset.
\newblock \emph{arXiv preprint arXiv:2105.07464}.

\bibitem[{French(1999)}]{french1999catastrophic}
Robert~M French. 1999.
\newblock Catastrophic forgetting in connectionist networks.
\newblock \emph{Trends in cognitive sciences}, 3(4):128--135.

\bibitem[{Fritzler et~al.(2019)Fritzler, Logacheva, and
  Kretov}]{fritzler2019few}
Alexander Fritzler, Varvara Logacheva, and Maksim Kretov. 2019.
\newblock Few-shot classification in named entity recognition task.
\newblock In \emph{Proceedings of the 34th ACM/SIGAPP Symposium on Applied
  Computing}, pages 993--1000.

\bibitem[{Gao et~al.(2020)Gao, Fisch, and Chen}]{gao2020making}
Tianyu Gao, Adam Fisch, and Danqi Chen. 2020.
\newblock Making pre-trained language models better few-shot learners.
\newblock \emph{arXiv preprint arXiv:2012.15723}.

\bibitem[{Geng et~al.(2019)Geng, Li, Li, Zhu, Jian, and
  Sun}]{geng2019induction}
Ruiying Geng, Binhua Li, Yongbin Li, Xiaodan Zhu, Ping Jian, and Jian Sun.
  2019.
\newblock Induction networks for few-shot text classification.
\newblock \emph{arXiv preprint arXiv:1902.10482}.

\bibitem[{Han et~al.(2018)Han, Zhu, Yu, Wang, Yao, Liu, and
  Sun}]{han2018fewrel}
Xu~Han, Hao Zhu, Pengfei Yu, Ziyun Wang, Yuan Yao, Zhiyuan Liu, and Maosong
  Sun. 2018.
\newblock Fewrel: A large-scale supervised few-shot relation classification
  dataset with state-of-the-art evaluation.
\newblock \emph{arXiv preprint arXiv:1810.10147}.

\bibitem[{Hou et~al.(2020)Hou, Che, Lai, Zhou, Liu, Liu, and Liu}]{hou2020few}
Yutai Hou, Wanxiang Che, Yongkui Lai, Zhihan Zhou, Yijia Liu, Han Liu, and Ting
  Liu. 2020.
\newblock Few-shot slot tagging with collapsed dependency transfer and
  label-enhanced task-adaptive projection network.
\newblock \emph{arXiv preprint arXiv:2006.05702}.

\bibitem[{Kulis et~al.(2013)}]{kulis2013metric}
Brian Kulis et~al. 2013.
\newblock Metric learning: A survey.
\newblock \emph{Foundations and Trends{\textregistered} in Machine Learning},
  5(4):287--364.

\bibitem[{Lample et~al.(2016)Lample, Ballesteros, Subramanian, Kawakami, and
  Dyer}]{lample2016neural}
Guillaume Lample, Miguel Ballesteros, Sandeep Subramanian, Kazuya Kawakami, and
  Chris Dyer. 2016.
\newblock Neural architectures for named entity recognition.
\newblock \emph{arXiv preprint arXiv:1603.01360}.

\bibitem[{Liu et~al.(2021)Liu, Yan, Lin, Han, and Sun}]{liu2021element}
Fangchao Liu, Lingyong Yan, Hongyu Lin, Xianpei Han, and Le~Sun. 2021.
\newblock Element intervention for open relation extraction.
\newblock \emph{arXiv preprint arXiv:2106.09558}.

\bibitem[{Loshchilov and Hutter(2017)}]{loshchilov2017decoupled}
Ilya Loshchilov and Frank Hutter. 2017.
\newblock Decoupled weight decay regularization.
\newblock \emph{arXiv preprint arXiv:1711.05101}.

\bibitem[{Ma et~al.(2022)Ma, Jiang, Wu, Zhao, and Lin}]{ma2022decomposed}
Tingting Ma, Huiqiang Jiang, Qianhui Wu, Tiejun Zhao, and Chin-Yew Lin. 2022.
\newblock Decomposed meta-learning for few-shot named entity recognition.
\newblock \emph{arXiv preprint arXiv:2204.05751}.

\bibitem[{Ma and Hovy(2016)}]{ma2016end}
Xuezhe Ma and Eduard Hovy. 2016.
\newblock End-to-end sequence labeling via bi-directional lstm-cnns-crf.
\newblock \emph{arXiv preprint arXiv:1603.01354}.

\bibitem[{Pearl(2009)}]{pearl2009causality}
Judea Pearl. 2009.
\newblock \emph{Causality}.
\newblock Cambridge university press.

\bibitem[{Peters et~al.(2017)Peters, Ammar, Bhagavatula, and
  Power}]{peters2017semi}
Matthew~E Peters, Waleed Ammar, Chandra Bhagavatula, and Russell Power. 2017.
\newblock Semi-supervised sequence tagging with bidirectional language models.
\newblock \emph{arXiv preprint arXiv:1705.00108}.

\bibitem[{Qi et~al.(2020)Qi, Niu, Huang, and Zhang}]{qi2020two}
Jiaxin Qi, Yulei Niu, Jianqiang Huang, and Hanwang Zhang. 2020.
\newblock Two causal principles for improving visual dialog.
\newblock In \emph{Proceedings of the IEEE/CVF conference on computer vision
  and pattern recognition}, pages 10860--10869.

\bibitem[{Schick and Sch{\"u}tze(2020)}]{schick2020s}
Timo Schick and Hinrich Sch{\"u}tze. 2020.
\newblock It's not just size that matters: Small language models are also
  few-shot learners.
\newblock \emph{arXiv preprint arXiv:2009.07118}.

\bibitem[{Snell et~al.(2017)Snell, Swersky, and Zemel}]{snell2017prototypical}
Jake Snell, Kevin Swersky, and Richard Zemel. 2017.
\newblock Prototypical networks for few-shot learning.
\newblock \emph{Advances in neural information processing systems}, 30.

\bibitem[{Tang et~al.(2020{\natexlab{a}})Tang, Huang, and Zhang}]{tang2020long}
Kaihua Tang, Jianqiang Huang, and Hanwang Zhang. 2020{\natexlab{a}}.
\newblock Long-tailed classification by keeping the good and removing the bad
  momentum causal effect.
\newblock \emph{Advances in Neural Information Processing Systems},
  33:1513--1524.

\bibitem[{Tang et~al.(2020{\natexlab{b}})Tang, Niu, Huang, Shi, and
  Zhang}]{tang2020unbiased}
Kaihua Tang, Yulei Niu, Jianqiang Huang, Jiaxin Shi, and Hanwang Zhang.
  2020{\natexlab{b}}.
\newblock Unbiased scene graph generation from biased training.
\newblock In \emph{Proceedings of the IEEE/CVF conference on computer vision
  and pattern recognition}, pages 3716--3725.

\bibitem[{Thrun(1995)}]{thrun1995learning}
Sebastian Thrun. 1995.
\newblock Is learning the n-th thing any easier than learning the first?
\newblock \emph{Advances in neural information processing systems}, 8.

\bibitem[{Tong et~al.(2021)Tong, Wang, Xu, Cao, Liu, Hou, and
  Li}]{tong2021learning}
Meihan Tong, Shuai Wang, Bin Xu, Yixin Cao, Minghui Liu, Lei Hou, and Juanzi
  Li. 2021.
\newblock Learning from miscellaneous other-class words for few-shot named
  entity recognition.
\newblock \emph{arXiv preprint arXiv:2106.15167}.

\bibitem[{Vinyals et~al.(2016)Vinyals, Blundell, Lillicrap, Wierstra
  et~al.}]{vinyals2016matching}
Oriol Vinyals, Charles Blundell, Timothy Lillicrap, Daan Wierstra, et~al. 2016.
\newblock Matching networks for one shot learning.
\newblock \emph{Advances in neural information processing systems}, 29.

\bibitem[{Wang et~al.(2021{\natexlab{a}})Wang, Xu, Liu, Zhou, Cao, Chang, and
  Sui}]{wang2021enhanced}
Peiyi Wang, Runxin Xu, Tianyu Liu, Qingyu Zhou, Yunbo Cao, Baobao Chang, and
  Zhifang Sui. 2021{\natexlab{a}}.
\newblock An enhanced span-based decomposition method for few-shot sequence
  labeling.
\newblock \emph{arXiv preprint arXiv:2109.13023}.

\bibitem[{Wang et~al.(2021{\natexlab{b}})Wang, Xun, Liu, Dai, Chang, and
  Sui}]{wang2021behind}
Peiyi Wang, Runxin Xun, Tianyu Liu, Damai Dai, Baobao Chang, and Zhifang Sui.
  2021{\natexlab{b}}.
\newblock Behind the scenes: An exploration of trigger biases problem in
  few-shot event classification.
\newblock In \emph{Proceedings of the 30th ACM International Conference on
  Information \& Knowledge Management}, pages 1969--1978.

\bibitem[{Wang et~al.(2020)Wang, Huang, Zhang, and Sun}]{wang2020visual}
Tan Wang, Jianqiang Huang, Hanwang Zhang, and Qianru Sun. 2020.
\newblock Visual commonsense r-cnn.
\newblock In \emph{Proceedings of the IEEE/CVF Conference on Computer Vision
  and Pattern Recognition}, pages 10760--10770.

\bibitem[{Wang et~al.(2021{\natexlab{c}})Wang, Chu, Zhang, and
  Gao}]{wang2021learning}
Yaqing Wang, Haoda Chu, Chao Zhang, and Jing Gao. 2021{\natexlab{c}}.
\newblock Learning from language description: Low-shot named entity recognition
  via decomposed framework.
\newblock \emph{arXiv preprint arXiv:2109.05357}.

\bibitem[{Yang and Katiyar(2020)}]{yang2020simple}
Yi~Yang and Arzoo Katiyar. 2020.
\newblock Simple and effective few-shot named entity recognition with
  structured nearest neighbor learning.
\newblock \emph{arXiv preprint arXiv:2010.02405}.

\bibitem[{Yu et~al.(2021)Yu, He, Zhang, Du, Pasupat, and Li}]{yu2021few}
Dian Yu, Luheng He, Yuan Zhang, Xinya Du, Panupong Pasupat, and Qi~Li. 2021.
\newblock Few-shot intent classification and slot filling with retrieved
  examples.
\newblock \emph{arXiv preprint arXiv:2104.05763}.

\bibitem[{Yue et~al.(2020)Yue, Zhang, Sun, and Hua}]{yue2020interventional}
Zhongqi Yue, Hanwang Zhang, Qianru Sun, and Xian-Sheng Hua. 2020.
\newblock Interventional few-shot learning.
\newblock \emph{Advances in neural information processing systems},
  33:2734--2746.

\bibitem[{Zeng et~al.(2020)Zeng, Li, Zhai, and Zhang}]{zeng2020counterfactual}
Xiangji Zeng, Yunliang Li, Yuchen Zhai, and Yin Zhang. 2020.
\newblock Counterfactual generator: A weakly-supervised method for named entity
  recognition.
\newblock In \emph{Proceedings of the 2020 Conference on Empirical Methods in
  Natural Language Processing (EMNLP)}, pages 7270--7280.

\bibitem[{Zhang et~al.(2020)Zhang, Zhang, Tang, Hua, and Sun}]{zhang2020causal}
Dong Zhang, Hanwang Zhang, Jinhui Tang, Xian-Sheng Hua, and Qianru Sun. 2020.
\newblock Causal intervention for weakly-supervised semantic segmentation.
\newblock \emph{Advances in Neural Information Processing Systems},
  33:655--666.

\bibitem[{Zhang et~al.(2021)Zhang, Lin, Han, and Sun}]{zhang2021biasing}
Wenkai Zhang, Hongyu Lin, Xianpei Han, and Le~Sun. 2021.
\newblock De-biasing distantly supervised named entity recognition via causal
  intervention.
\newblock \emph{arXiv preprint arXiv:2106.09233}.

\end{thebibliography}
\bibliographystyle{acl_natbib}

\appendix
\clearpage

\section{Formula Appendix}
\label{sec:appendix}

\begin{small}
\begin{align}
    P(Y=y|do(T=t))= \sum_{E}&P(Y=y|do(T=t),E=e)
    \notag
    \\&P(E=e|do(T=t))
\end{align}
\end{small}

\begin{small}
\begin{align}
    P(Y=y|do&(T=t))=\sum_{E}P(Y=y|do(T=t),
    \notag
    \\ &do(E=e))P(E=e|do(T=t))
\end{align}
\end{small}

\begin{small}
\begin{align}
    P(Y=y|do&(T=t))=\sum_{E}P(Y=y|do(T=t),
    \notag
    \\ &do(E=e))P(E=e|T=t)
\end{align}
\end{small}

\begin{small}
\begin{align}
    P(Y=y&|do(T=t))=
    \notag
    \\ &\sum_{E}P(Y=y|do(E=e))P(E=e|T=t)
\end{align}
\end{small}

\begin{small}
\begin{align}
    P&(Y=y|do(T=t))=\sum_{t'}\sum_{E}P(Y=y|do(E=e),
    \notag
    \\ &T=t')P(T=t'|do(E=e))P(E=e|T=t)
\end{align}
\end{small}

\begin{small}
\begin{align}
    P(Y=y|&do(T=t))=\sum_{t'}\sum_{E}P(Y=y|E=e,T=t')
    \notag
    \\ &P(T=t'|do(E=e))P(E=e|T=t)
\end{align}
\end{small}

\begin{small}
\begin{align}
    P(Y=y|do&(T=t))=\sum_{t'}\sum_{E}P(Y=y|do(E=e),
    \notag
    \\ &T=t')P(T=t')P(E=e|T=t)
\end{align}
\end{small}

\begin{small}
\begin{align}
    P(Y=y&|do(T=t))=\sum_{E}P(E=e|T=t)
    \notag
    \\ &\sum_{t'}P(Y=y|E=e,T=t')P(T=t')
\end{align}
\end{small}

\section{Baseline Appendix}

ProtoBERT uses BERT to get the vector representation of each token, and then averages all vectors of the same type as the the class representation according to the prototype network. Finally, calculate the distance between each category representation and query, and judge based on the nearest class.

NNShot gets the feature representation for each token and calculates the distance between the query and each representation. Finally, the class is judged based on the nearest distance.

StructShot adds an additional Viterbi decoder to the NNShot.

CONTAINER also uses BERT, and additionally uses contrast learning and Gaussian embedding to get the representation of each token. Then, fine-tuning on the support set and inference using the nearest neighbor method.

ESD uses inter and cross-span attention based on prototypes to get span representations. Also, it constructs multi-prototypes for O label.

Decomposed Meta-Learning considers the few shot as a sequence labeling problem. MAML is used to initialize the model parameters, and meanwhile uses MAML-Protonet to find the optimal embedding space for entity recognition.

\section{Experiments Appendix}

\textbf{Parameter Setting} We use BERT-base-uncased. Besides, more parameter setting details are shown in the table.

\begin{table}[h]
\normalsize
  \centering
  \begin{spacing}{1.1}
    \begin{tabular}{cc}
    \toprule
    Name & Value  \\
    \cmidrule(lr){1-2}
    Batch\_size & 20  \\
    Max\_length & 32  \\
    Learning rate & 1e-4  \\
    Embedding dimension & 768  \\
    Dropout & 0.1  \\
    \bottomrule
    \end{tabular}%
    \end{spacing}
  \caption{The hyperparameters of experiments}
  \label{tab:addlabel}%
\end{table}%

\end{document}